\pdfoutput=1

\documentclass[11pt]{article}

\usepackage[preprint]{acl}

\usepackage{times}
\usepackage{latexsym}

\usepackage[T1]{fontenc}

\usepackage[utf8]{inputenc}

\usepackage{microtype}

\usepackage{inconsolata}

\usepackage{graphicx}

\usepackage{subfiles}
\usepackage{amsmath}
\usepackage{graphicx}

\title{Depth $F_1$: Improving Evaluation of Cross-Domain \\ Text Classification by Measuring Semantic Generalizability}

\author{Parker Seegmiller, Joseph Gatto, and Sarah Masud Preum \\
  Department of Computer Science \\
  Dartmouth College \\
  Hanover, NH, USA \\
  \texttt{pkseeg.gr@dartmouth.edu} \\}

\begin{document}
\maketitle
\begin{abstract}
Recent evaluations of cross-domain text classification models aim to measure the ability of a model to obtain domain-invariant performance in a target domain given labeled samples in a source domain. The primary strategy for this evaluation relies on assumed differences between source domain samples and target domain samples in benchmark datasets. This evaluation strategy fails to account for the similarity between source and target domains, and may mask when models fail to transfer learning to specific target samples which are highly dissimilar from the source domain. We introduce Depth $F_1$, a novel cross-domain text classification performance metric. Designed to be complementary to existing classification metrics such as $F_1$, Depth $F_1$  measures how well a model performs on target samples which are dissimilar from the source domain\footnote{Depth $F_1$ code can be found at \url{https://github.com/pkseeg/df1/}}. We motivate this metric using standard cross-domain text classification datasets and benchmark several recent cross-domain text classification models, with the goal of enabling in-depth evaluation of the semantic generalizability of cross-domain text classification models.
\end{abstract}

\section{Introduction}

Cross-domain text classification is the task of predicting labels for texts in a \textit{target} domain, given only a labeled set of texts in a \textit{source} domain. There exist several approaches to this task, including kernel-based approaches \cite{muandet2013domain}, contrastive learning \cite{tan2022domain}, key-value memory adaptation \cite{jia2023memory}, and most recently the use of large language models \cite{jia2022prompt, long2023adapt}.

\begin{figure}[htbp]
    \centerline{
        \includegraphics[width=\columnwidth]{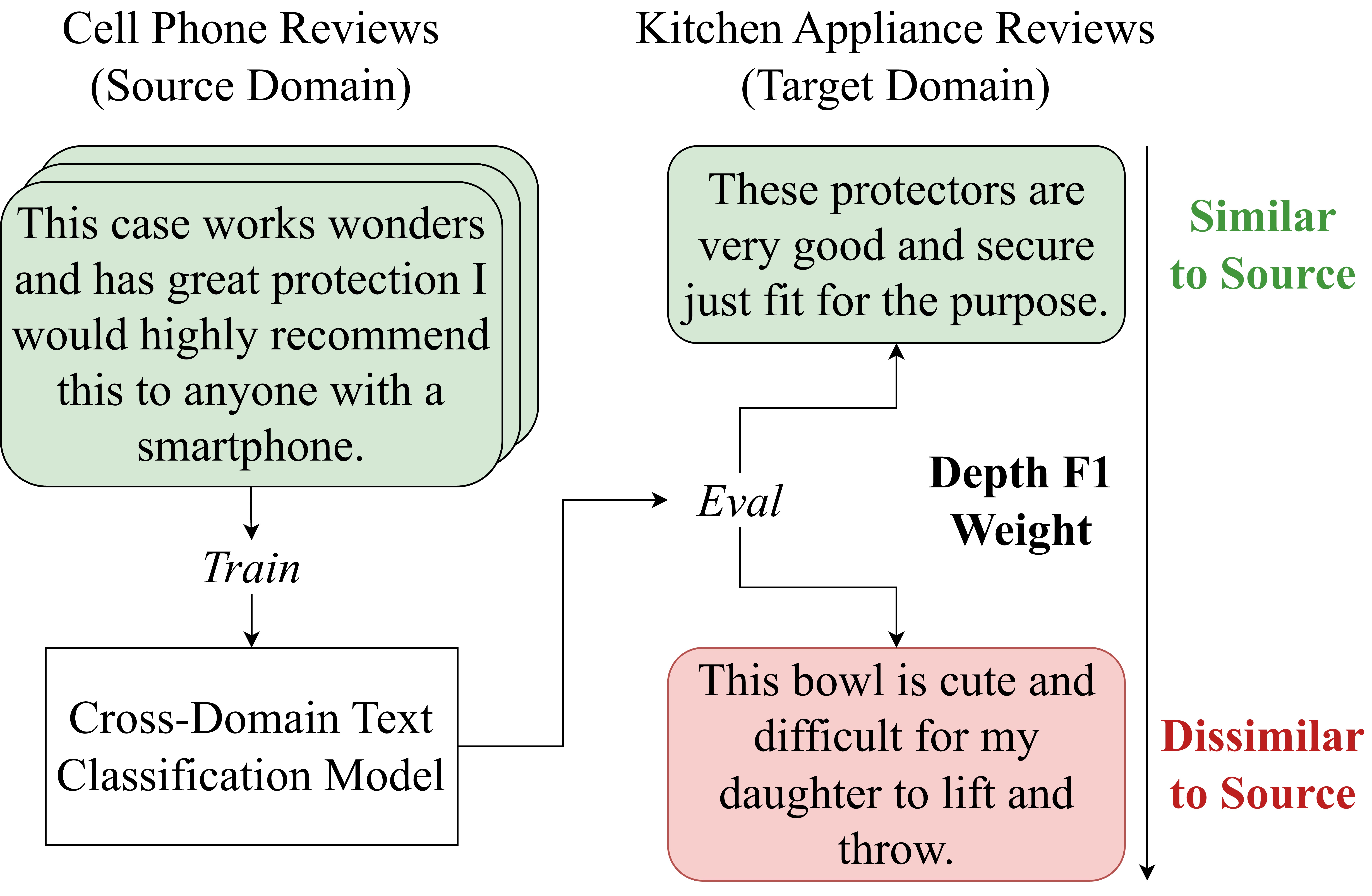}
    }
    \caption{Depth $F_1$ ($DF_1$) is a cross-domain text classification metric designed to measure a model's semantic generalizability. Both kitchen appliance reviews have positive sentiments. Still, the one highlighted in red is more challenging to classify due to its dissimilarity to samples in the source domain cell phone reviews. $DF_1$ re-weights target samples by dissimilarity to the source domain, enabling a more in-depth evaluation of model performance. Detailed discussion of these examples can be found in Appendix \ref{app:sim_examples}.}
    \label{fig:preview}
\end{figure}

Effective cross-domain text classification evaluation plays a key role in developing better cross-domain text classification models \cite{moreo2021lost}. To evaluate text classification models in a cross-domain setting, a model is trained on a source domain $S$ and evaluated using a dissimilar target domain $T$ \cite{muandet2013domain}. Predicted target labels are then compared with ground truth labels using standard classification evaluation metrics such as $F_1$. This is commonly performed several times, in a k-fold cross-domain evaluation strategy \cite{jia2023memory}. In order to measure the ability of the model to transfer its learning across domains, this evaluation strategy relies on dissimilarity between source and target domains in a given dataset. When source and target domains are dissimilar, good performance indicates a model's ability to learn domain-generalizable features from source texts and labels \cite{gulrajani2020search}. 

However, the dissimilarity between source and target domains is hardly measured. Even if the dissimilarity is measured, it is measured on a macro scale by comparing aspects of the overall distributions of source $S$ and target $T$ \cite{kour2022measuring, bu2022characterizing}. Thus, existing cross-domain text classification evaluations often fail to properly measure the ability of a model to transfer to specific samples in the target domain that are dissimilar from those in the source domain. We hypothesize that in cross-domain text classification, \textbf{it is challenging for a model to transfer knowledge learned from a source domain to target domain samples that are semantically dissimilar from the source domain (i.e., source-dissimilar)}. Failing to distinguish between these samples in evaluation leads to a misleading estimation of the generalizability of a model, as it is possible to label only source-similar target samples correctly.

Overconfidence in the ability of a model to generalize has potentially risky implications in safety-critical domains \cite{yang2023test}, e.g., biology, cybersecurity, law, medicine, or politics. Consider the case of a disease classification model \cite{li-etal-2020-analysis}, trained on an open-source set of clinical notes such as i2b2 \cite{patrick2010high} (i.e., source domain). To evaluate the performance of this model in a cross-domain setting, researchers use a set of labeled clinical notes from their local clinic (target domain). These labeled target domain clinical notes are largely similar to i2b2 notes, but contain a small amount of highly dissimilar notes mentioning rare symptoms and diseases (e.g., a rare type of cancer). While the model may achieve good overall performance in this setting according to $F_1$, using $F_1$ by itself fails to measure the performance of the disease classification model on the small amount of clinical notes that are dissimilar to the i2b2 source domain. Since these highly dissimilar clinical notes contain rare disease symptoms, this masking of poor performance on source-dissimilar clinical notes can be dangerous for downstream applications. Deploying this model could mean that patients with rare, complex symptoms might be misdiagnosed and not receive proper medical care. Masking a model's performance on specific source-dissimilar target domain samples is especially relevant for data from underrepresented and marginalized populations \cite{gianfrancesco2018potential, mehrabi2021survey}.



To address this gap in cross-domain text classification evaluation, we introduce Depth-$F_1$ ($DF_1$), a novel $F_1$-based metric that assigns more performance weight to target samples that are dissimilar to source domain samples. Designed to be complementary to existing classification metrics such as $F_1$ score, $DF_1$ aims to quantify the \textit{semantic generalizability} of cross-domain text classification models. Our specific contributions are as follows. (i) We develop the mathematical framework for $DF_1$, which utilizes a statistical depth function for measuring instance-level differences in source and target domains \cite{seegmiller2023statistical}. (ii) Through extensive experiments using modern transfer learning classification models on two benchmark cross-domain text classification datasets, we highlight the need for $DF_1$ by finding and exploring instances where poor model performance on source-dissimilar target domain samples is masked by current evaluation strategies. We demonstrate the effectiveness of $DF_1$ in providing a comprehensive evaluation of model performance on source-dissimilar target domain samples. 

\section{Related Works}



\subsection{Cross-Domain Text Classification Benchmarks}
Most cross-domain text classification works follow a benchmark-based evaluation strategy. In benchmark-based evaluation, the standard $F_1$ metric, i.e., the harmonic mean between precision and recall, is used to evaluate models on some held-out test set from the target domain. This is commonly done in leave-one-out cross-domain evaluation \cite{gulrajani2020search}. \cite{ben2022pada} use several multi-source/single-domain evaluation scenarios, including rumor detection and natural language inference (NLI). A common evaluation benchmark dataset is the Multi-Genre NLI (MultiNLI) dataset, which contains natural language inference samples in ten distinct genres of English \cite{ben2022pada, williams2017broad}. \cite{jia2023memory} also use MultiNLI, as well as sentiment analysis of Amazon customer reviews \cite{blitzer2007biographies}. The Amazon customer reviews are another common evaluation benchmark in cross-domain text classification model evaluation, in which sentiment analysis is the classification task \cite{tan2022domain, jia2022prompt, long2023adapt}.  

Our proposed $DF_1$ metric enables a more in-depth analysis of models on these benchmark datasets, or any text classification dataset with a source and target domain. $DF_1$ enables researchers to measure the semantic generalizability of models by shifting evaluation focus to target samples which are highly dissimilar to source domains in benchmark datasets.

\subsection{Measuring Distances Between Corpora}
In evaluating models, some studies measure the distance between source and target datasets, for example by measuring the maximum mean discrepancy \cite{borgwardt2006integrating} between source and target domains \cite{yan2019weighted, kour2022measuring, bu2022characterizing}. In these evaluations, distance is measured between the source and target domains generally, disjoint from model evaluation.  A common approach to measuring differences in source and target domains is to compare $F_1$ scores of a given model with $F_1$ scores of an oracle model trained on the target domain \cite{moreo2021lost}. The disparity of results between the baseline and oracle models on the target domain is used to indicate the disparity between the two datasets. There are several other ways of measuring the distance between two textual corpora that operate on a distributional level, either via semantics-encoded document vectors such as SBERT \cite{reimers-2019-sentence-bert} or token distributions \cite{kour2022measuring}. 

Embeddings-based distributional metrics such as MAUVE, which estimates the gap between two corpora by using divergence frontiers \cite{pillutla2021mauve}, are used to obtain a single measure of the distance between two text datasets. These distance metrics are sometimes used in cross-domain text classification evaluation to show source-target dissimilarity \cite{yan2019weighted}. 

In contrast to these, $DF_1$ assigns each sample in the target domain a weight indicating its dissimilarity from the source domain. The ability to obtain a different measure of semantic dissimilarity from a source domain for each text sample in a target domain enables a more granular examination of model performance and generalizability in that domain \cite{seegmiller2023statistical}. This type of examination is complementary to existing distance measures and enables more thorough error analysis when evaluating cross-domain text classification models \cite{wu-etal-2019-errudite}.

\section{A Novel Evaluation Strategy for Cross-Domain Text Classification}
\label{sec:methods}

We first introduce Depth-$F_1$ ($DF_1$), a performance metric that scores cross-domain text classification models on their semantic generalizability, i.e., the ability to generalize to target domain samples that are dissimilar from the source domain. We then define $\lambda$, the single hyperparameter for $DF_1$, which creates increasingly semantically-dissimilar and challenging evaluation subsets of the target domain. $DF_1$ measures how well a cross-domain text classification model transfers to dissimilar samples.

\subsection{Depth Weights $w_i$}


Our goal is to define, for each sample in the target domain $t_i \in T$, a \textit{weight} $w_i$ which approximates how dissimilar $t_i$ is from the source samples on which a model is trained. We then use $w_i$ to weight target domain samples in the $F_1$ calculation. 

We first adopt the definition of transformer-based text embedding (TTE) depth from \cite{seegmiller2023statistical}. That is, given a set of source domain samples, $S$, and a set of target domain samples, $T$, TTE depth assigns a score to each sample in the target domain $t_i \in T$ indicating how well $t_i$ represents $S$. 

TTE depth encodes source domain and target domain samples using a cosine-based text encoder model to get a set of source sample embeddings $X_S$ and a set of target sample embeddings $X_T$. TTE depth then scores each target sample embedding $x_{t_i} \in X_T$ according to

\begin{equation}
\label{eq:depth}
    D_\delta(x_{t_i},X_S) := 2 - E_{x_S}[\delta(x_{t_i},H)]
\end{equation}

where $H \sim X_S$ is a random variable with uniform distribution over $X_S$, and $\delta$ is cosine distance. In this work, we utilize SBERT\footnote{sentence-transformers/all-MiniLM-L6-v2} \cite{reimers-2019-sentence-bert} as the text encoder. However, any standard cosine-based text encoder model can be used in practice \cite{almarwani-diab-2021-discrete, li2023angle}. We discuss how the SBERT model is specifically applicable for our Depth $F_1$ definition in Appendix \ref{app:details_sbert}. We next define the \textit{source domain median} $s_0 \in S$, as in \cite{seegmiller2023statistical}, to be the sample in the source domain with maximal depth.

\begin{equation}
\label{eq:median}
    \max\limits_{x_s \in X_S} D_\delta(x_s, X_S) = D_\delta(x_{s_0}, X_S)
\end{equation}

For shorthand, we use $D_\delta(a)$ to denote $D_\delta(x_{a},X_S)$, the TTE depth of a sample $a$ (in either $T$ or $S$) with respect to the source domain $S$.

We are now ready to define a weight for approximating the dissimilarity of a sample in the target domain $t_i \in T$ from the source domain $S$. Using the TTE depth of the source domain median, we define a weight $w_i$ for each sample in the target domain $t_i \in T$.

\begin{equation}
\label{eq:weight}
    w_i := \frac{D_\delta(s_0) - D_\delta(t_i)}{\sum\limits_{t_j \in T}(D_\delta(s_0) - D_\delta(t_j))}
\end{equation}

The depth of each sample in the target domain $D_\delta(t_i)$ measures how dissimilar $t_i$ is, on average, from the source domain samples in embedding space. Comparing the depth of each sample in the target domain $t_i \in T$ with the depth of the source domain median $s_0$ measures how far this similarity is from the "average similarity" of the source domain samples, enabling comparison of weights across pairs of \{source, target\} domains. Normalizing this measure ensures that all weights sum to $1$. As $DF_1$  is meant to be a complementary measure to $F_1$, scaling weights allows greater interpretability between the two metrics.

\subsection{Depth $F_1$ ($DF_1$)}

Using weights $w_0, ..., w_N$ for samples in the target domain $t_0, ..., t_N \in T$, we define $DF_1$ over a set of discrete ground truth labels $y_0, ..., y_N \in Y$ and a set of discrete predicted labels $\hat{y}_0, ..., \hat{y}_N \in \hat{Y}$. $DF_1$ counts the depth-weighted true positives among the predicted labels simply by defining $true(\hat{y}_i)$ to be

\begin{equation}
\label{eq:true}
true(\hat{y}_i) = 
\begin{cases}
  w_i & \hat{y}_i = y_i \\
  0 & \text{otherwise}
\end{cases}
\end{equation}

and counting depth-weighted true positives by

\begin{equation}
\label{eq:tp}
DTP = \sum_{\hat{y}_i \in Y} true(\hat{y}_i)
\end{equation}

Using an analogous construction for depth-weighted false positives $DFP$ and false negatives $DFN$, $DF_1$ is defined as

\begin{equation}
\label{eq:df1}
DF_1 = \frac{2DTP}{2DTP + DFP + DFN}
\end{equation} 

\subsection{$\lambda$ Dissimilarity}
We might reasonably expect a cross-domain text classification model to generalize to samples in the target domain that are similar to those from the source domain used for training. As such, a more challenging benchmark evaluation would isolate those target domain samples which are sufficiently dissimilar from the source domain, and measure how well a model performs on only these dissimilar target samples. To form this more challenging benchmark of semantic generalizability for a given source and target domain, we wish to exclude some amount of samples from our measurement that are not sufficiently dissimilar to the source domain. We define the $\lambda$ hyperparameter, which subsets the set of evaluation samples in the target domain $T$ to include only those samples $t_i$ with greater weight $w_i$ than a given percentile $\lambda$. Formally, given a percentile $0 < \lambda < 100$, we find the sample in the target domain $t_\lambda$ which has $\lambda$th percentile TTE depth among all samples in the target domain, w.r.t. the source domain. We then replace the evaluation target set $T$ with the set $T_\lambda$. 

\begin{equation}
\label{eq:tlambda}
T_\lambda = \{t_i \in T, D_\delta(t_i) < D_\delta(t_\lambda) \}
\end{equation} 

Substituting $T_\lambda$ for $T$ in Equations \ref{eq:weight} onward, we can calculate $DF_{1_\lambda}$ for any $\lambda$ percentile $0 < \lambda < 100$. Higher values of $\lambda$ indicate more challenging evaluation subsets of the target domain. As $\lambda$ increases, the task of transferring knowledge from the source domain to the subset $T_{\lambda}$ bridges a more semantically dissimilar gap, enabling a more in-depth measurement of the semantic generalizability of a cross-domain text classification model.
\section{Benchmark Cross-Domain Text Classification Data}
\label{sec:data}

We now describe two standard benchmark datasets used in cross-domain evaluation. We quantify how source and target domains differ semantically in these datasets. We then demonstrate how $F_1$ may offer a misleading estimation of model generalizability in these datasets, and how $DF_1$ sheds light on this discrepancy by demonstrating the behavior of models overfitting to source-similar texts.


\subsection{Benchmark Data}
\label{sec:benchmark}

\begin{table*}[!ht]
  \centering
    \begin{tabular}{lllllc}
        Pairing & Source & Task & Source & Target & $Q$\\\hline
        SiS-1 & Single & SA & Cell Phone & Baby Product & 0.34\\
        SiS-2 & Single & SA & Yelp & Cell Phone & 0.33\\
        MuS-1 & Multi & SA & Cell Phone, Baby Product, Yelp & IMDB & 0.35\\
        MuS-2 & Multi & SA & Baby Product, IMDB, Cell Phone & Yelp & 0.39\\\hline
        SiS-3 & Single & NLI & Travel & Government & 0.34\\
        SiS-4 & Single & NLI & Slate & Telephone & 0.42\\
        MuS-3 & Multi & NLI & Travel, Slate, Government, Telephone & Fiction & 0.48\\
        MuS-4 & Multi & NLI & Telephone, Slate, Government, Fiction & Travel & 0.38\\\hline
    \end{tabular}
    \caption{Source and target domain pairings used in experiments, along with $Q$ statistics for measuring domain similarity. Each source domain contains 5,000 samples, except for SiS-3 and Sis-4 which contain 2,000 each. Each target domain contains 1,000 samples. A lower $Q$ indicates that these two domains are more dissimilar in general, according to TTE depth.}
    \label{tab:dataq}
\end{table*}

\textbf{Sentiment Analysis (SA):} Following recent work in cross-domain sentiment analysis \cite{he2018adaptive, xue2020improving}, we draw our sentiment analysis cross-domain evaluation scenarios from four domains of user reviews. The IMDB and Yelp domains \cite{tang2015learning} contain movie and business reviews, respectively. Following \cite{he2018adaptive}, the Cell phone and Baby product domains contain product reviews from Amazon \cite{blitzer2007biographies}. We consider the challenging multi-class classification setting, as reviews from the Yelp, Cell phone, and Baby product domains are labeled 1-5. IMDB reviews are originally labeled 1-10; we scale these to 1-5 for simplicity.

\textbf{Natural Language Inference (NLI):} We also consider the multi-genre natural language inference (NLI) dataset \cite{williams2018broad}, where the task is to predict the directional relation between a pair of sentences (entailment, contradiction, neither). A sampled version of this dataset was introduced in \cite{ben2022pada} and is commonly used in cross-domain text classification evaluation \cite{jia2022prompt, jia2023memory}. This dataset covers five domains: fiction, government, slate, telephone, and travel. In our experiments, NLI sentence pairs are separated by a newline character and jointly encoded, as in \cite{ben2022pada}.

We consider both single-source (SiS) and multi-source (MuS) scenarios. Given that across the two datasets there exist 41 total \{domain, target\} pairings, we randomly select two single-source and two multi-source pairings from each dataset as listed in Table \ref{tab:dataq}. Following \cite{ben2022pada}, we downsample the size of the datasets in these pairings to emphasize the need for a cross-domain algorithm. For each pairing, we randomly sample 5K samples from the source domain(s). As the single-source domain NLI datasets contain fewer than 5K samples, in these instances, we randomly sample 2K samples from each source domain. Each target domain contains 1K randomly selected samples.


\subsection{Data Investigation with TTE Depth}
To further motivate the need for $DF_1$, we demonstrate that not all \{source, target\} domain pairings are equally dissimilar. Using $F_1$ alone in these scenarios may be masking the ability, or inability, of a cross-domain text classification model to generalize to dissimilar samples in a target domain. In each domain pairing in Table \ref{tab:dataq}, we take the TTE depth scores of each source sample, $D_\delta(s_i) \forall s_i \in S$, and each target sample $D_\delta(t_i) \forall t_i \in T$. These scores indicate how similar each text sample is to the source domain. A higher score indicates that the sample is more similar to the source domain, and a lower score indicates that the sample is more dissimilar to the source domain. 

This dissimilarity between a source domain $S$ and target domain $T$, measured by TTE depth scores, can be quantified using the $Q$ statistic: 

\begin{equation}
\label{eq:q}
Q(S,T) = Pr[D_\delta(X,S) \le D_\delta(Y,T)]
\end{equation}

where $X \sim S$ and $Y \sim T$ are independent random variables with uniform distributions over $S$ and $T$. This statistic measures how dissimilar two corpora are in general, by calculating the empirical probability that a randomly selected sample from the source domain has a lower TTE depth score than a randomly selected sample from the target domain. The lower the $Q$ measure between a source domain $S$ and target domain $T$, the more dissimilar the corpora. $Q$ statistic estimates are given in Table \ref{tab:dataq}.

\begin{figure}[htbp]
    \centerline{
        \includegraphics[width=\columnwidth]{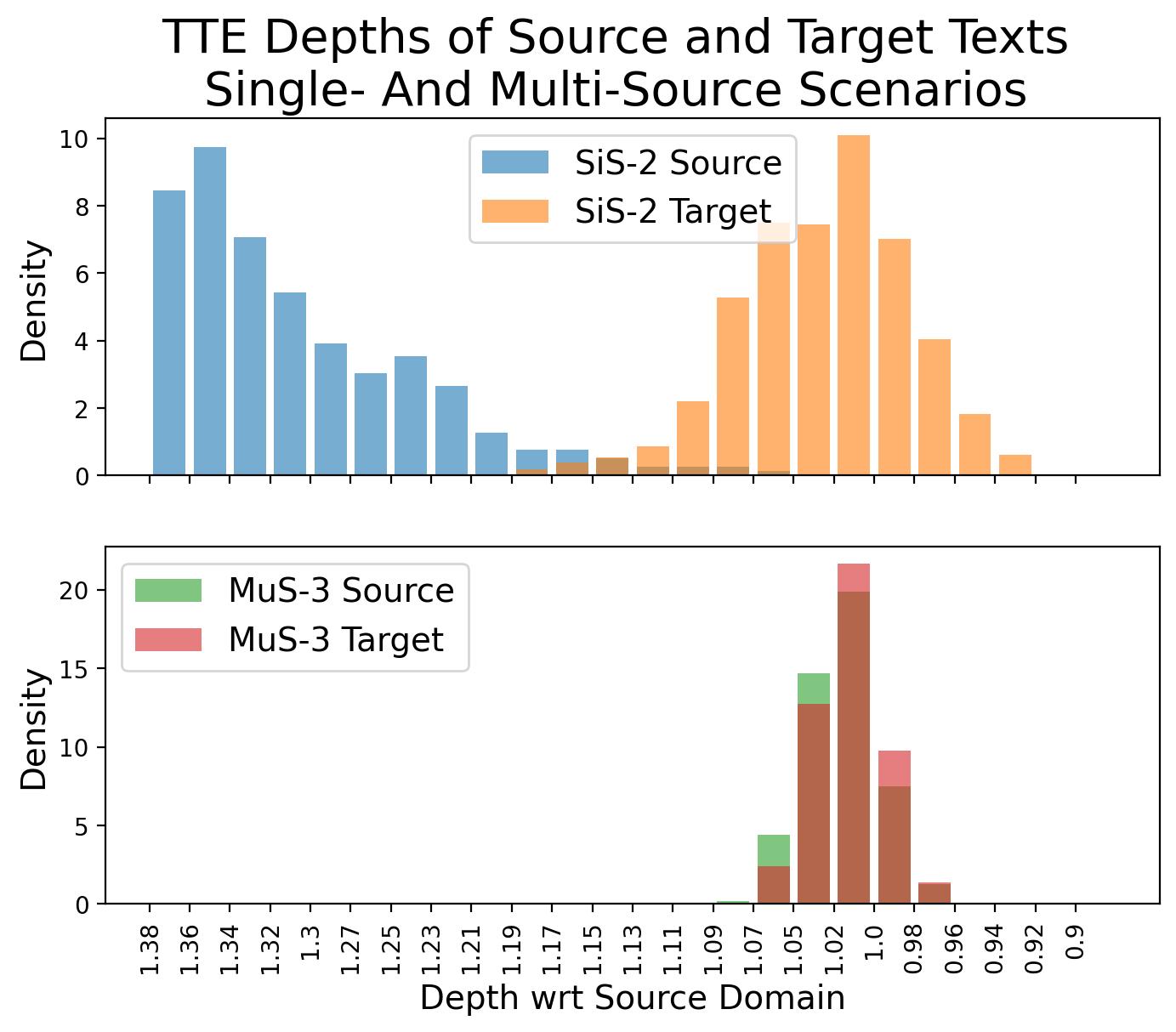}
    }
    \caption{TTE depth scores of source and target samples in SiS-2 and MuS-3 pairings, with respect to source samples. The separated nature of the SiS-2 source and target domains indicates that the two domains are highly semantically dissimilar. However, MuS-3 is more semantically overlapping (brown bars resulting from overlap in the distributions indicated by red and green). This discrepancy highlights that not all cross-domain text classification datasets present equally challenging tasks.}
    \label{fig:depth}
\end{figure}

Utilizing the SiS-2 and MuS-3 pairings, as these are the pairings with the least and most similarity between source and target domains, we plot the source and target domain TTE depth scores in Figure \ref{fig:depth}. We see that in SiS-2, TTE depth scores of the source samples are higher overall and much more separated from their target domain, and they present a more dissimilar, more challenging cross-domain setting. The inclusion of multiple sources in the MuS-3 pairing, on the other hand, leads to lower source sample depth scores and less dissimilarity overall. The disparity between the source and target domain similarities in these two pairings highlights the need for $DF_1$. While standard cross-domain text classification evaluation with $F_1$ alone would treat performance in these two pairings equally, $DF_1$ more heavily weights dissimilar samples in the target domain and rewards model performance on these more challenging samples. 

Examples of ``easy'' and ``challenging'' target domain samples that may occur in evaluating cross-domain text classification can be seen in Figure \ref{fig:preview}. An in-depth investigation into such samples is provided in Appendix \ref{app:sim_examples}.

\subsection{Scope of $F_1$ for Cross-Domain Text Classification}
\label{sec:illustration}

In this section, we utilize the SiS-1 domain pairing to illustrate the problem that arises when using only $F_1$ for measuring cross-domain performance of text classification models, and highlight how $DF_1$ helps to solve this problem. We define two \textit{demonstration models} -- i.e. toy models designed to highlight a specific model behavior -- for this illustration. Demonstration model $A$ can classify all target texts with an equal probability of success. Demonstration model $B$ is more able (i.e., with a higher probability of success) to classify target texts which are similar to the source domain, and is less able to classify target texts which are dissimilar to the source domain.

\begin{figure}[htbp]
    \centerline{
        \includegraphics[width=\columnwidth]{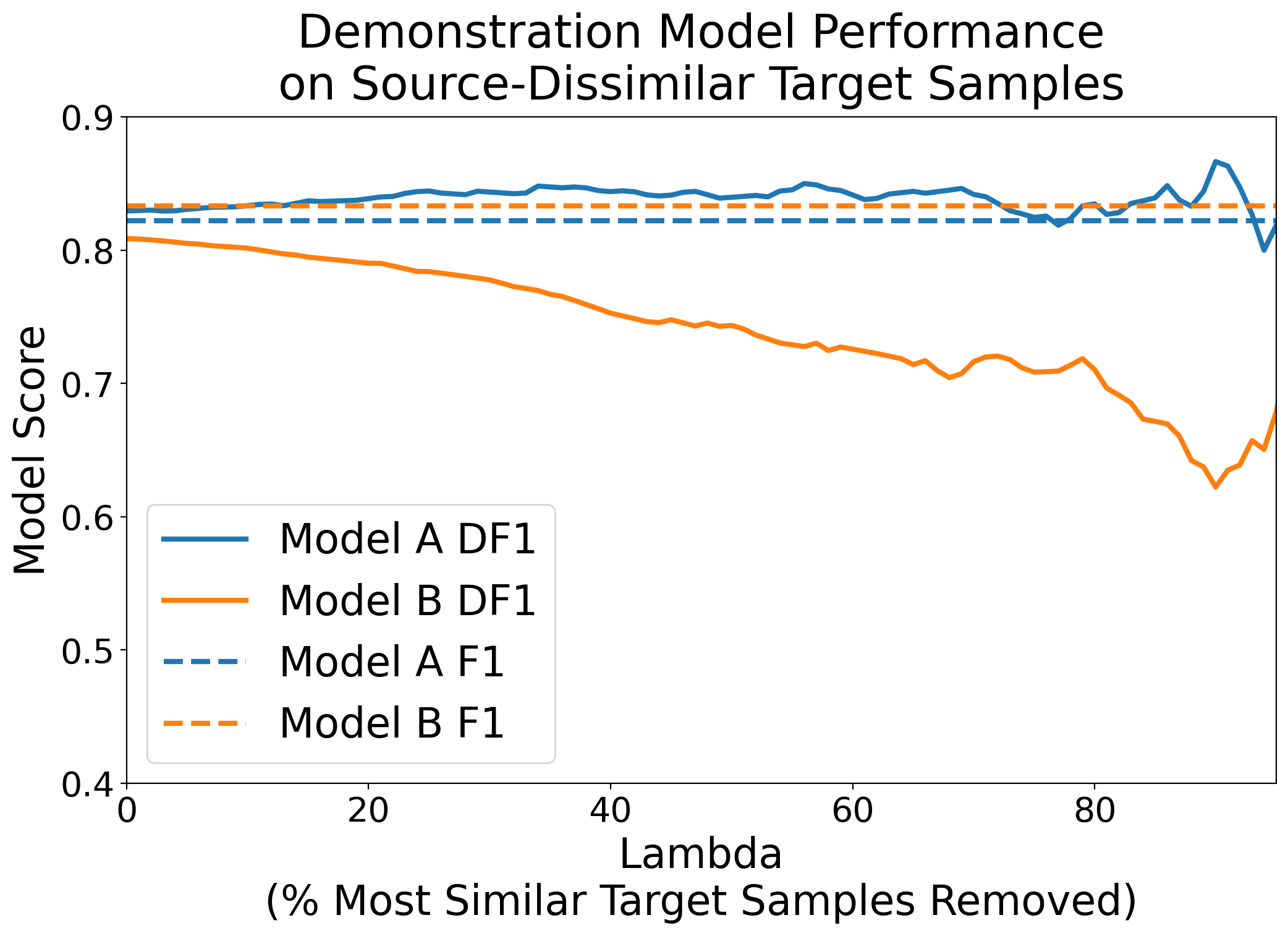}
    }
    \caption{$F_1$ and Depth $F_1$ ($DF_1$) results of the two demonstration models $A$ and $B$ on the SiS-1 cross-domain text classification \{source, domain\} pairing. While models $A$ and $B$ have nearly identical $F_1$ scores, they differ significantly in $DF_1$ scores, and that difference increases with $\lambda$ as more domain-similar target texts are removed from the target domain evaluation set.}
    \label{fig:problem}
\end{figure}

Formally, for a given set of target domain texts $t_i$ at the $\lambda_i$th percentile TTE depth from the source domain, we consider two text classification models $A(t_i, \lambda_i, \gamma_A)$ and $B(t_i, \lambda_i, \gamma_B)$. Model $A$ is equally likely to predict a correct target label for text $t_i$ with probability $\gamma_A$. Model $B$ selects a correct label for text $t_i$ with probability $\gamma_B$, where $\gamma_B$ decreases incrementally as $\lambda_i$ increases. We evaluate models $A$ and $B$ on the SiS-1 sentiment analysis domain pairing, where cell phone reviews are the source domain and baby product reviews are the target domain. We set $\gamma_A = 0.83$ and $\gamma_B = 0.95$. Results are shown in Figure \ref{fig:problem}. 

While model $A$ and model $B$ have similar $F_1$ scores at $0.822$ and $0.833$, respectively, we see that the performances of the two models are drastically different when it comes to source dissimilar texts. The $DF_1$ scores are $0.829$ and $0.809$ for models $A$ and $B$, respectively, and as $\lambda$ increases the $DF_1$ scores of $B$ decrease significantly. Once $\lambda$ reaches $50$, $DF_{1_{\lambda = 50}} = 0.744$. The drop in model $B$ performance indicates an undesirable model trait; model $B$ is unable to transfer knowledge to target texts that are dissimilar from the source domain. In machine learning terms, model $B$ \textit{overfits}, and performs worse on samples dissimilar to the source domain compared with model $A$. Where standard cross-domain text classification evaluation obscures this overfitting on source-similar texts, $DF_1$ measures this model behavior.
\section{Experiments}


We demonstrate how using $DF_1$ to measure cross-domain text classification performance enables explicit evaluation of semantic generalizability in modern deep-learning-based cross-domain text classification models by benchmarking several models with $DF_1$ on all 8 domain pairings in Table \ref{tab:dataq}. We then provide a detailed analysis of three types of model behavior under $DF_1$.

\subsection{Benchmarking Models Using $DF_1$}
\label{sec:results}

We benchmark 4 recent cross-domain text classification models using $DF_1$. We consider both \textit{fine-tuned} models which are trained on source data directly, and \textit{prompt-based} models that attempt to utilize vast amounts of linguistic knowledge stored in pre-trained language models to label target texts.

\textbf{SBERT+LR} \cite{reimers-2019-sentence-bert}: As a simple baseline, we train a logistic regression model that takes as input SBERT  document representations of source and target texts.

\textbf{MSCL} \cite{tan2022domain}: Tan et al. introduce a Memory-based Supervised Contrastive Learning (MSCL) model, which learns domain-invariant representations by utilizing supervised contrastive learning with a memory-storing queue. 

\textbf{RAG} \cite{long2023adapt}: Long et al. use as a baseline a retrieval-augmented generation (RAG) strategy, which uses measures of vector similarity to select in-context learning demonstrations from source domain texts for each target domain text at inference time.

\textbf{DAICL} \cite{long2023adapt}: Long et al. propose a Domain-Adapative In-Context Learning (DAICL) model, which is a retrieval-augmented text generation strategy during which similar unlabeled target domain texts are used in conjunction with labeled source examples during training, then target texts are labeled using retrieval-augmented generation to select in-context learning demonstrations from source domain texts.

\begin{figure*}[htbp]
    \centerline{
        \includegraphics[width=1.9\columnwidth]{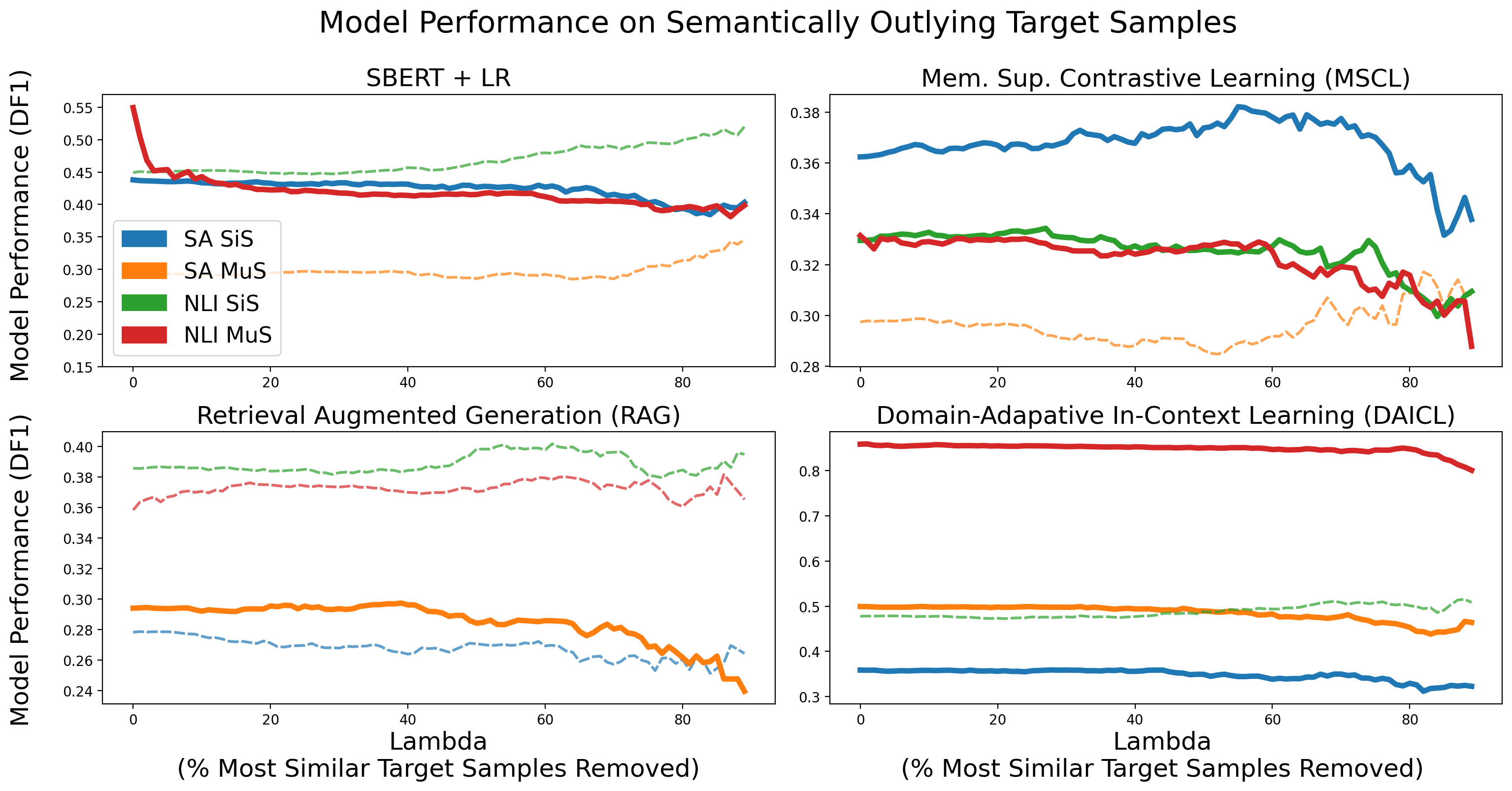}
    }
    \caption{Evaluation of cross-domain text classification models using $F_1$ and Depth $F_1$ ($DF_1$). We present both micro-average $F_1$ scores and micro-average $DF_1$ scores from $\lambda = 0$ to $\lambda = 90$, i.e., the percentage of most similar target texts that are not considered in the evaluation. Each result is averaged across two \{source, target\} domain pairings for both the sentiment analysis (SA) and natural language inference (NLI) tasks in both the single-source (SiS) and multi-source (MuS) scenarios.  Model performance that decreases as $\lambda$ increases is indicated with solid lines, highlighting overfitting on source-similar texts. We give these results, along with corresponding F1 scores, in tabular format in Tables \ref{tab:experiments_sa} and \ref{tab:experiments_nli} of Appendix \ref{app:investigate}.}
    
    \label{fig:results}
\end{figure*}

Model training details are found in Appendix \ref{app:training_details}. We benchmark each model on each domain pairing. $DF_1$ scores are shown in Figure \ref{fig:results}. We find that the model behavior of \textit{semantic overfitting}, illustrated in Section \ref{sec:illustration}, is pervasive. This undesirable model trait occurs across both tasks, in both single-source and multi-source scenarios, for both fine-tuned and prompt-based models. 

\subsection{Examining Model Behavior}
We discuss three examples of cross-domain text classification model performance under $DF_1$, as seen in Figure \ref{fig:results}: (i) $DF_1$ decreases as $\lambda$ increases, (ii) $DF_1$ is consistent as $\lambda$ increases, and (iii) $DF_1$ increases as $\lambda$ increases.

\subsubsection{$DF_1$ Decreases as $\lambda$ Increases}


We highlight the RAG model performance on the MuS-2 domain pairing. In this domain setting, models are trained on baby product reviews, IMDB reviews, and cell phone reviews, and evaluates using Yelp reviews. In the case of RAG, rather than training, for each target domain review the model selects similar reviews from a pool of source domain reviews to be used in in-context learning. Interestingly, model performance according to $DF_1$ drops significantly as $\lambda$ increases -- in other words, as evaluation is shifted to more source-dissimilar target domain reviews. This behavior, which the demonstration models highlight in Section \ref{sec:illustration}, can be seen in Table \ref{tab:experiments_sa} of Appendix \ref{app:investigate}.

As an example of reviews that are considered source dissimilar according to depth, take this review of the Smith Center from the Yelp target domain (labeled \textit{extremely positive}): ``Yes, Virginia, there is culture in Las Vegas and it begins at the Smith Center.'' In the context of Yelp, this review is easily recognizable as having positive sentiment. However, in the context of baby products, cell phones, and movies, this review is more challenging. The RAG model performs poorly in transferring knowledge, either from pre-training or from selected contextual samples, to these types of source-dissimilar reviews. 

\subsubsection{$DF_1$ Consistent as $\lambda$ Increases}

In contrast to the RAG model, which performs more poorly on source-dissimilar target domain samples on the MuS-2 domain pairing, the MSCL model remains relatively consistent as $\lambda$ increases. This consistent performance can be seen in Table \ref{tab:experiments_sa} of Appendix \ref{app:investigate}. This is a desirable model behavior, as it indicates that the MSCL model is more semantically generalizable, as it is able to maintain consistent performance as evaluation shifts to more source-dissimilar target domain reviews.

\subsubsection{$DF_1$ Increases as $\lambda$ Increases}

As seen in Figure \ref{fig:results} and Table \ref{tab:experiments_nli} of Appendix  \ref{app:investigate}, the SBERTLR model maintains mostly consistent performance across the multi-source sentiment analysis domain pairing of training on baby product, Yelp, and cell phone reviews, and evaluating on IMDB reviews (MuS-1). However, as $\lambda$ reaches 90 and only 10\% of the target domain samples remain in the evaluation set, performance according to $DF_1$ increases slightly. While this behavior can be partially attributed to chance variation, it can also be due to the characteristics of the source dissimilar target domain samples in this dataset. An investigation into the target domain IMDB reviews with low $DF_1$ weights shows that, while dissimilar to the source domain reviews, these samples are easier in general for the sentiment analysis task. While our evidence shows that highly source-dissimilar target domain samples are more challenging in general, this is not always the case. This behavior of performance increase alongside $\lambda$ happens less frequently than other behaviors. So, we leave further investigation of this behavior for future work.

\section{Relevance to Other NLP Tasks}
While we focus on text classification in this work, our approach may be useful for \textit{any} cross-domain task with a suitable statistical depth function. Evaluation of models for other NLP tasks such as machine translation \cite{sutskever2014sequence}, question answering \cite{xu2023critical}, summarization \cite{zou2021low}, or information extraction \cite{tong2022docee} in the cross-domain setting may benefit from investigation into source-dissimilar target domain samples using TTE depth. Future work may investigate the evaluation of semantic generalizability for these tasks.
\section{Conclusion}

As NLP models are being used more widely across various tasks, driven in part by the development of large language models \cite{yang2024harnessing}, it is becoming increasingly important to identify and assess the limitations and blind spots of these models. We have shown that the current evaluation of cross-domain text classification models is limited in its measurement of semantic generalizability, i.e., generalizability with respect to semantically dissimilar source and target domain samples. This motivates the development of a novel cross-domain text classification performance metric, Depth $F_1$, which measures how well a model performs on target samples that are dissimilar from the source domain. Depth $F_1$ offers valuable new insight into modern approaches to cross-domain text classification. We believe that the $DF_1$ metric will enable a more in-depth evaluation of cross-domain generalization of text classification models.


\section{Limitations}
Our definition of $DF_1$ relies on the ability of SBERT embeddings to encode semantic similarity between pairs of texts in cross-domain text classification datasets. While using SBERT embeddings in conjunction with cosine distance is a common strategy for measuring similarity between two texts \cite{gatto2023text, chen-etal-2022-generate, gao-etal-2021-simcse}, there may be other strategies which are better suited for specific source and target domains in various scenarios. Our $DF_1$ definition allows for this flexibility in downstream cross-domain text classification evaluations. Our definition of $DF_1$ also relies on the TTE depth value of the pre-defined source domain median. While the depth of the median is useful for our definition, the median is not meant to be the definitive representative text from the source domain. However, we leave investigation of $DF_1$ using different similarity measures for a specific domain, and the effect of source domain median selection, for future work.  


\bibliography{custom}

\appendix

\section{Examples of Similarity and Dissimilarity in Cross-Domain Datasets}
\label{app:sim_examples}
To highlight the types of ``easy'' and ``challenging'' target domain samples that may occur in evaluating cross-domain text classification, we highlight samples in SiS-1, where the source domain is cell phone reviews and the target domain is baby product reviews. A truncated version of these samples can be seen in Figure \ref{fig:preview}. Consider the source median from the cell phone reviews: ``This works wonders and has great protection I would highly recommend this to anyone with a smartphone.'' This source text is labeled a 5 for positive sentiment. A target text from the baby product reviews which has a higher TTE depth score (1.23) with respect to the cell phone reviews is ``These protectors are very good and secure just fit for the purpose, the quantity and the quality are good too. I highly recommend this item.'' This target text is also labeled a 5 for positive sentiment. Note that the ``protector'' being reviewed here is similar to the cell phone cases you might find in the source domain. We call target domain samples like this \textit{source-similar} in that, measured by TTE depth, they have high average semantic similarity to the source domain. 

Note that there is a semantic difference between source-similar target samples and target samples like the following: ``This bowl is cute and difficult for my daughter to lift and throw, however if your high chair has a smaller tray this bowl will not fit.'' This target text is labeled a 4 for moderately positive sentiment, yet it  is \textit{source-dissimilar} in that it has a much lower TTE depth score (0.94) with respect to the source domain cell phone reviews. We hypothesize that in cross-domain text classification tasks such as sentiment analysis, it is easier for a model to transfer knowledge learned from source domain samples to source-similar target samples, as opposed to source-dissimilar target samples.

The $DF_1$ metric reduces the amount of performance weight given to models for performing well on the source-similar target samples in the overlapping areas between source and target TTE depth scores in Figure \ref{fig:depth}, and increases the performance weight for source-dissimilar samples in the tail of the target domain.

\section{Model Details}
\label{app:details}

\subsection{Sentence BERT}
\label{app:details_sbert}
As described in Section \ref{sec:methods}, we utilize SBERT as the cosine-based text encoder model in our deployment of Depth $F_1$. SBERT is primarily trained in a contrastive manner according to the triplet objective function

\begin{equation}
\label{eq:contrastiveloss}
    \max(||x_a - x_p|| - ||x_a - x_n|| + \epsilon, 0)
\end{equation}

where $x_p$ and $x_n$ are network embeddings of positive and negative text samples, respectively, from labeled datasets, and $x_n$ is the SBERT network embedding of an anchor text sample. The stated goal of the SBERT model is to produce embeddings which are semantically meaningful when compared with cosine similarity. The objective function in Equation \ref{eq:contrastiveloss} learns embeddings such that $x_p$ is at least $\epsilon$ closer to $x_a$ than $x_n$, and aims to encode semantic relationships in embedding space via distance between embeddings. Hence, a low cosine distance between a pair of SBERT embeddings indicates the two are semantically similar, whereas a high cosine distance indicates the pair is semantically dissimilar.

\subsection{Benchmark Training Details}
\label{app:training_details}

Details of each model evaluated in Section \ref{sec:results} are given here for reproducibility. All models were trained and evaluated using Python 3.1. 

\textbf{SBERT+LR} \cite{reimers-2019-sentence-bert}: The SBERT+LR model utilizes the sentence-transformers library\footnote{all-MiniLM-L6-v2} and the sklearn library, training an sklearn logistic regression model using default parameters on the source domain.

\textbf{MSCL} \cite{tan2022domain}: The MSCL model was trained and evaluated using the source code from the original paper\footnote{https://github.com/tonytan48/MSCL}. As in the original work, we use the roberta-large\footnote{https://huggingface.co/FacebookAI/roberta-large} model in the transformers library, trained for 5 epochs on a single A100 gpu with a learning rate of 1e-5. Including these, all hyperparameters follow the original work.

\textbf{RAG} \cite{long2023adapt}: For the RAG model, we utilize cosine similarity over SBERT embeddings as our vector store. We utilize the top 3 most similar source domain samples as demonstrations for each target domain sample, and prompt Llama 2\footnote{https://huggingface.co/meta-llama/Llama-2-7b-hf} to assign a label.

\textbf{DAICL} \cite{long2023adapt}: The DAICL model, which is an augmented retrieval-augmented text generation strategy, is similar to the RAG model. As opposed to Llama 1 which was utilized in the intial paper, we again utilize Llama 2, this time first fine-tuning the model on a single A100 gpu for 3 epochs with a learning rate of 3e-4. We also utilize cosine similarity over SBERT embeddings as our vector store.

\section{Tabular Results of Model Performance using $DF_1$}
\label{app:investigate}
Results from Section \ref{sec:results} can be seen in Tables \ref{tab:experiments_sa} and \ref{tab:experiments_nli} for the sentiment analysis (SA) and natural language inference (NLI) domain pairings, respectively. We present both micro-average $F_1$ scores and micro-average $DF_1$ scores across $\lambda = \{0, 25, 50, 75, 90\}$, across both single-source (SiS) and multi-source (MuS) domain pairings.

\begin{table*}[!ht]
  \centering
    \begin{tabular}{cccccccc}\hline
    Pairing & Model &$F_1$ &$DF_{1_{\lambda=0}}$ &$DF_{1_{\lambda=25}}$ &$DF_{1_{\lambda=50}}$ &$DF_{1_{\lambda=75}}$ &$DF_{1_{\lambda=90}}$ \\\hline
    \textbf{SiS-1} &\textbf{SBERTLR} &\textbf{0.585} &\textbf{0.585} &\textbf{0.584} &\textbf{0.579} &\textbf{0.548} &\textbf{0.517} \\
    \textbf{SiS-1} &\textbf{MSCL} &\textbf{0.507} &\textbf{0.509} &\textbf{0.507} &\textbf{0.522} &\textbf{0.506} &\textbf{0.457} \\
    SiS-1 &RAG &0.332 &0.330 &0.324 &0.325 &0.317 &0.330 \\
    \textbf{SiS-1} &\textbf{DAICL} &\textbf{0.348} &\textbf{0.338} &\textbf{0.331} &\textbf{0.318} &\textbf{0.303} &\textbf{0.273} \\\hline
    \textbf{SiS-2} &\textbf{SBERTLR} &\textbf{0.297} &\textbf{0.291} &\textbf{0.279} &\textbf{0.274} &\textbf{0.257} &\textbf{0.261} \\
    SiS-2 &MSCL &0.212 &0.216 &0.224 &0.226 &0.234 &0.244 \\
    \textbf{SiS-2} &\textbf{RAG} &\textbf{0.229} &\textbf{0.226} &\textbf{0.215} &\textbf{0.217} &\textbf{0.200} &\textbf{0.211} \\
    \textbf{SiS-2} &\textbf{DAICL} &\textbf{0.380} &\textbf{0.379} &\textbf{0.383} &\textbf{0.381} &\textbf{0.371} &\textbf{0.339} \\\hline
    MuS-1 &SBERTLR &0.291 &0.293 &0.298 &0.283 &0.305 &0.336 \\
    MuS-1 &MSCL &0.293 &0.293 &0.291 &0.273 &0.280 &0.313 \\
    \textbf{MuS-1} &\textbf{RAG} &\textbf{0.312} &\textbf{0.315} &\textbf{0.319} &\textbf{0.313} &\textbf{0.310} &\textbf{0.306} \\
    \textbf{MuS-1} &\textbf{DAICL} &\textbf{0.423} &\textbf{0.418} &\textbf{0.420} &\textbf{0.404} &\textbf{0.388} &\textbf{0.369} \\\hline
    MuS-2 &SBERTLR &0.290 &0.292 &0.296 &0.288 &0.304 &0.363 \\
    MuS-2 &MSCL &0.304 &0.302 &0.299 &0.300 &0.317 &0.296 \\
    \textbf{MuS-2} &\textbf{RAG} &\textbf{0.282} &\textbf{0.274} &\textbf{0.272} &\textbf{0.255} &\textbf{0.228} &\textbf{0.171} \\
    \textbf{MuS-2} &\textbf{DAICL} &\textbf{0.586} &\textbf{0.581} &\textbf{0.578} &\textbf{0.575} &\textbf{0.536} &\textbf{0.562} \\
    \hline
    \end{tabular}
    \caption{Tabular results of cross-domain text classification models on the sentiment analysis (SA) pairings, both single-source (SiS) and multi-source (MuS) using $F_1$ and Depth $F_1$ ($DF_1$). We present both micro-average $F_1$ scores and micro-average $DF_1$ scores across $\lambda = \{0, 25, 50, 75, 90\}$, i.e. the percentage of most-similar target texts which are not considered in evaluation.  In bold, we see that model performance decreases as $\lambda$ increases in several models across all four domain pairings, indicating overfitting on source-similar texts.}
    \label{tab:experiments_sa}
\end{table*}

\begin{table*}[!ht]
  \centering
    \begin{tabular}{cccccccc}\hline
    Pairing & Model &$F_1$ &$DF_{1_{\lambda=0}}$ &$DF_{1_{\lambda=25}}$ &$DF_{1_{\lambda=50}}$ &$DF_{1_{\lambda=75}}$ &$DF_{1_{\lambda=90}}$ \\\hline
    SiS-3 &SBERTLR &0.453 &0.457 &0.449 &0.461 &0.489 &0.495 \\
    SiS-3 &MSCL &0.352 &0.353 &0.357 &0.347 &0.356 &0.370 \\
    \textbf{SiS-3} &\textbf{RAG} &\textbf{0.382} &\textbf{0.385} &\textbf{0.384} &\textbf{0.403} &\textbf{0.394} &\textbf{0.359} \\
    SiS-3 &DAICL &0.528 &0.531 &0.529 &0.545 &0.572 &0.588 \\\hline
    SiS-4 &SBERTLR &0.415 &0.442 &0.447 &0.465 &0.502 &0.549 \\
    \textbf{SiS-4} &\textbf{MSCL} &\textbf{0.312} &\textbf{0.306} &\textbf{0.310} &\textbf{0.306} &\textbf{0.298} &\textbf{0.216} \\
    SiS-4 &RAG &0.378 &0.386 &0.386 &0.394 &0.368 &0.403 \\
    SiS-4 &DAICL &0.406 &0.425 &0.424 &0.432 &0.443 &0.404 \\\hline
    MuS-3 &SBERTLR &0.383 &0.677 &0.422 &0.406 &0.393 &0.399 \\
    \textbf{MuS-3} &\textbf{MSCL} &\textbf{0.339} &\textbf{0.353} &\textbf{0.348} &\textbf{0.353} &\textbf{0.336} &\textbf{0.318} \\
    MuS-3 &RAG &0.369 &0.338 &0.365 &0.364 &0.365 &0.376 \\
    MuS-3 &DAICL &0.827 &0.846 &0.838 &0.837 &0.845 &0.819 \\\hline
    MuS-4 &SBERTLR &0.426 &0.422 &0.421 &0.424 &0.408 &0.432 \\
    \textbf{MuS-4} &\textbf{MSCL} &\textbf{0.320} &\textbf{0.310} &\textbf{0.311} &\textbf{0.303} &\textbf{0.285} &\textbf{0.257} \\
    \textbf{MuS-4} &\textbf{RAG} &\textbf{0.391} &\textbf{0.379} &\textbf{0.384} &\textbf{0.377} &\textbf{0.391} &\textbf{0.357} \\
    \textbf{MuS-4} &\textbf{DAICL} &\textbf{0.878} &\textbf{0.871} &\textbf{0.872} &\textbf{0.863} &\textbf{0.847} &\textbf{0.783} \\
    \hline
    \end{tabular}
    \caption{Tabular results of cross-domain text classification models on the natural language inference (NLI) pairings, both single-source (SiS) and multi-source (MuS) using $F_1$ and Depth $F_1$ ($DF_1$). We present both micro-average $F_1$ scores and micro-average $DF_1$ scores across $\lambda = \{0, 25, 50, 75, 90\}$, i.e. the percentage of most-similar target texts which are not considered in the evaluation.  In bold, we see that model performance decreases as $\lambda$ increases in several models across all four domain pairings, indicating overfitting on source-similar texts.}
    \label{tab:experiments_nli}
\end{table*}

\end{document}